# Automated Estimation of Impact Time, Impact Location, and Shuttlecock Speed in Badminton Smashes Using Event Cameras

Yudai Washida [1]*, Yuto Kase [1], Kai Ishibe [1], Ryoma Yasuda [1] and Sakiko Hashimoto [1]

[1]*MIZUNO Corporation, 1-12-35 Nanko-Kita, Suminoe-ku, Osaka-shi, Osaka 559-8510, Japan;*

*Corresponding author
Yudai Washida: ywashida@mizuno.co.jp

ORCID iDs
Yudai Washida: 0009-0007-3466-7451
Yuto Kase: 0009-0002-1512-0331
Kai Ishibe: 0009-0004-0470-9540
Ryoma Yasuda: 0009-0000-2620-1896
Sakiko Hashimoto: 0009-0002-9536-5354

**Abstract**

Quantifying impact phenomena in badminton smashes is important for evaluating both athletic performance and equipment; however, conventional measurement systems involve trade-offs between temporal resolution, data efficiency, and preparation effort. This study proposes a measurement method using two synchronized event cameras to automatically estimate impact time, impact location on the racket face, and post-impact shuttlecock speed in an integrated manner within the same trial.

The swing interval was detected from event rate statistics, impact time was estimated from the shuttlecock trajectory inflection in the lateral-view event data, impact location was determined by ellipse fitting to the racket face in the rear-view event image, and shuttlecock speed was calculated in the sagittal plane. To validate the proposed method, Bland–Altman analysis was performed against a high-speed camera-based reference method using 125 smash trials from five players. Impact time and shuttlecock speed were estimated in all 124 analyzable trials, and impact location was estimated in 93.5% (116/124). The bias (95% CI) for impact time, medio-lateral impact location, longitudinal impact location, and shuttlecock speed were 1.84 ms (1.45 to 2.23), 3.45 mm (2.18 to 4.72), −1.92 mm (−2.97 to −0.88), and −1.00 m/s (−2.46 to 0.46), respectively. No proportional bias was observed for any metric. These results suggest that the proposed method can serve as a useful tool for integrated assessment of badminton smash performance and equipment in practical settings.

## 1. Introduction

In badminton smashes, impact conditions strongly influence shot outcome [1, 2]. Quantifying the impact phenomenon is therefore important for evaluating both athletic performance and equipment. However, several challenges remain in the measurement and quantitative assessment of impact phenomena.

First, smash-related metrics such as impact location and shuttlecock trajectory vary considerably across trials, within individuals, and between individuals [2, 3]; a sufficient number of trials is thus required for reliable characterization. Second, because contact duration is only 1.0–1.4 ms [1, 4], evaluating impact location on the racket face at this time scale requires high temporal resolution. Third, a systematic understanding of impact phenomena requires integrated evaluation of these metrics rather than individual assessment. Finally, reducing the effort required for measurement preparation and post-processing is also important from a practical standpoint.

High-speed cameras and optical three-dimensional motion capture systems have been used to acquire racket and shuttlecock data in badminton [1, 2, 5]. High-speed cameras are well suited for capturing fast phenomena; however, the large data volumes generated at high resolution and high frame rates can constrain multi-trial measurements. Motion capture systems provide accurate coordinate acquisition [2]; however, sampling frequencies are typically limited to approximately 1,000 Hz, constraining temporal resolution. In addition, the preparation and post-processing burden—including marker attachment, calibration, and labeling—is substantial, and attaching markers to the shuttlecock may affect its flight characteristics.

Event cameras can potentially address these challenges. Unlike conventional frame-based cameras that acquire images at fixed intervals, event cameras asynchronously record per-pixel brightness changes [6]. This operating principle enables microsecond-level temporal resolution with low motion blur. The output consists solely of the timestamp t, pixel coordinates (x, y), and polarity (positive or negative; $p \in \{+1, -1\}$) of each brightness change; data from static background regions are inherently suppressed, providing high data efficiency well suited for multi-trial measurements [6].

In the sports domain, event cameras have been applied to ball spin estimation and high-speed ball motion analysis [7, 8]. We have previously reported automatic impact location estimation in tennis [9] and estimation of impact time and post-impact flight characteristics in badminton [10] using event cameras. However, no study has yet implemented and validated an automated processing pipeline that associates impact location, impact time, and post-impact flight within the same trial across multiple trials.

This study proposes a measurement method that uses event cameras to automatically estimate impact time, impact location on the racket face, and post-impact shuttlecock speed in an integrated manner within the same trial during badminton smashes. To validate the proposed method, its agreement with a high-speed camera-based reference method was evaluated using Bland–Altman analysis.

## 2. Method

### 2.1. Overview

Two synchronized event cameras were used to automatically estimate (i) the swing interval, (ii) impact time, (iii) impact location on the racket face, and (iv) post-impact shuttlecock tracking (Fig. 1b). The impact time estimated from the lateral view was propagated to the rear view, where only events near that time were processed, thereby stabilizing and accelerating impact location estimation. High-speed cameras recorded the same trials for validation.

### 2.2. Experimental Setup

#### 2.2.1. Participants

Five healthy badminton players (four males, one female; all right-handed) participated. Their age, height, body mass, and playing experience were 30.2 ± 4.9 years (range: 23–36), 169.8 ± 7.7 cm, 68.0 ± 10.8 kg, and 17.0 ± 7.9 years, respectively. All participants received a written explanation of the study and provided written informed consent. The study was approved by the Mizuno Corporation Research Ethics Committee (Ref. No. 20260408-475).

#### 2.2.2. Environment and Layout

Experiments were conducted in an indoor gymnasium with sufficient illuminance (1,215 lx). To capture the impact and shuttlecock flight, two viewpoints—lateral and rear—were established, each equipped with a closely positioned pair of an event camera (EVK4; Prophesee, France; 1,280 × 720 pixels) and a high-speed camera (Fastcam Mini AX100; Photron, Japan) (Fig. 1a). The high-speed camera resolution and frame rate were set as 1,024 × 560 pixels at 2,000 fps for the lateral view and 1,024 × 880 pixels at 5,000 fps for the rear view, with an exposure time of 0.2 ms for both. The cameras in each viewpoint were positioned to share approximately the same field of view. To facilitate event detection, a dark curtain was placed behind the measurement volume, and a racket with a white frame and black strings was used (Fig. 1c).

Each participant performed 25 maximal smashes toward a target placed in front of them. Shuttlecocks were fed by an automatic shuttlecock launcher to ensure a consistent high-lob trajectory. The two event cameras were time-synchronized via a synchronization signal that established a common time base. To synchronize the event cameras with the high-speed cameras, a reference trigger was manually generated for each trial. Because this trigger provided a common temporal reference rather than the physical impact time, the present study primarily evaluated the inter-method difference in impact time.

Spatial calibration was performed using a target of known dimensions to establish the correspondence between pixel coordinates and physical length. The resulting spatial resolutions of the event cameras were 3.47 mm/pixel for the lateral view and approximately 1.86 mm/pixel for the rear view, the latter estimated per trial from the racket outline.

### 2.3. Proposed Algorithm

#### 2.3.1. Swing Interval Detection and Impact Time Estimation

A reference impact time is required to associate impact location and shuttlecock speed within the same trial. Based on our previous work [9, 10], the swing interval was first detected from the event data and then used to estimate impact time (Fig. 1b). During a swing, the high-speed motion of the racket increases the event rate. Because events are generated asynchronously, they were aggregated into 0.5 ms bins, each termed an event packet. The mean and variance of the event rate were computed over a sliding 100 ms window of consecutive event packets; the swing interval onset was defined as the moment at which both statistics exceeded their respective thresholds, and the interval was set to the subsequent 100 ms (Fig. 2a, b). Event data within the detected swing interval were then extracted (Fig. 2c) and projected onto the x–t plane, where the shuttlecock trajectory is represented along the time axis and exhibits an abrupt change around impact (Fig. 2e). The time at which this inflection occurs in the x-coordinate trajectory was defined as the impact time. Coupling swing interval detection [9] with impact time estimation within a known interval [10] enabled fully automated processing.

#### 2.3.2. Shuttlecock Speed Estimation

Immediately after impact, the shuttlecock undergoes posture change and flipping, making its behavior irregular [4]. To avoid this transient phase, event data were extracted at 30 ms and 32 ms after impact (each accumulated over 1 ms), the position of the shuttlecock tip was identified in each image, and the shuttlecock speed in the sagittal plane was calculated [10] (Fig. 4d).

#### 2.3.3. Impact Location Estimation

Using the impact time estimated from the lateral view, event data from the rear view were retrieved to estimate impact location. Data were processed sequentially as event packets at 0.5 ms intervals; when the estimated impact time fell within the time range of a packet, the events in that packet were accumulated to form an image. The temporal resolution of the event packets was equivalent to 2,000 fps.

The impact location estimation procedure, originally developed for tennis [9], was adapted for badminton. Whereas tennis defines the impact location as the ball center, this study defined it as the shuttlecock tip. First, positive-polarity regions were extracted from the rear-view event image, and regions with a sufficient area in the upper portion of the image were selected as candidates (Fig. 3a). Next, the lateral extent of the positive-polarity regions was used to separate the racket-face region from the shaft region (Fig. 3b). An ellipse was then fitted to the racket-face region to estimate its geometric shape (Fig. 3c). Among the positive-polarity events within the estimated racket face, the lowest point—corresponding to the tip of the downward-falling shuttlecock—was identified as the shuttlecock tip. A RANSAC line fit was applied to the shaft region, and the resulting shaft orientation was used to correct the tilt of the fitted ellipse. Finally, a racket-face coordinate system was defined with the corrected ellipse center as the origin, the minor axis as the medio-lateral axis (u-axis), and the major axis as the longitudinal axis (v-axis); the shuttlecock tip coordinates in this system were taken as the impact location. The inner dimensions of the racket frame were 184 mm (medio-lateral) and 240 mm (longitudinal) (Fig. 1c).

Trials in which the shuttlecock tip fell outside the corrected ellipse, or in which the estimated shaft orientation deviated more than ±60° from vertical, were classified as failed trials and excluded from the accuracy evaluation. This separation allowed the success rate of the algorithm and the inter-method agreement for successful trials to be evaluated independently.

## 2.4. Validation

### 2.4.1. High-Speed Camera-Based Measurement

High-speed camera images were manually digitized to obtain impact location and shuttlecock speed as reference measurements. Impact time was defined as the frame in which the shuttlecock tip completely separated from the string bed in the lateral view and was determined by visual inspection (Fig. 4c). Impact location was calculated from the synchronized rear-view image at the identified impact time (Fig. 4a): the inner contour of the racket frame was annotated as an ellipse and the shuttlecock tip as a point using VGG Image Annotator (VIA; University of Oxford), and the impact location was computed in the racket-face coordinate system from the ellipse parameters and tip coordinates. Shuttlecock speed was calculated from the lateral-view images at 30 ms and 32 ms after impact by annotating the shuttlecock tip in each frame and computing the sagittal-plane speed (Fig. 4c).

### 2.4.2. Statistical analysis

Statistical analyses were performed in Python using the statsmodels library and custom code. Bland–Altman analysis [11] was used to assess the agreement between the proposed method and the high-speed camera-based reference method. The bias (mean difference) and 95% limits of agreement (LoA) were calculated for impact time, medio-lateral and longitudinal impact location, and shuttlecock speed. Impact time was defined as the difference from the detection time of the reference trigger signal. The bias, LoA, and their 95% confidence intervals were calculated following the method of Zou [12], which accounts for repeated measurements within participants. Pearson's correlation coefficient between the mean and the difference was computed to check for proportional bias. Of the 125 total trials, one trial in which the reference trigger was not detected was excluded from the analysis.

**3. Results**

The proposed method successfully estimated impact time in all 124 trials. Among these, impact location was estimated in 93.5% of trials (116/124), with per-participant success rates of 91.7% (22/24), 100% (25/25), 92.0% (23/25), 88.0% (22/25), and 96.0% (24/25). Shuttlecock speed was estimated in all 124 trials.

The mean ± standard deviation of impact time, medio-lateral impact location, longitudinal impact location, and shuttlecock speed estimated by the proposed method were −29.39 ± 52.96 ms, −24.43 ± 19.77 mm, −17.05 ± 29.47 mm, and 51.66 ± 4.73 m/s, respectively. The corresponding values obtained from the high-speed camera reference method were −31.23 ± 52.94 ms, −27.89 ± 19.61 mm, −15.13 ± 29.69 mm, and 52.64 ± 4.78 m/s, respectively.

Representative estimation results obtained by the proposed method (EV) and the high-speed camera-based method (HS) are shown in Fig. 4, showing generally consistent estimates. To quantify this agreement, Bland–Altman plots are shown in Fig. 5.

For impact time (Fig. 5a), the bias was 1.84 ms (95% CI: 1.45 to 2.23), and the LoA ranged from −0.84 ms (95% CI: −1.48 to −0.44) to 4.52 ms (95% CI: 4.12 to 5.16). For medio-lateral impact location (Fig. 5c), the bias was 3.45 mm (95% CI: 2.18 to 4.72), and the LoA ranged from −3.35 mm (95% CI: −5.59 to −2.18) to 10.24 mm (95% CI: 9.07 to 12.49). For longitudinal impact location (Fig. 5d), the bias was −1.92 mm (95% CI: −2.97 to −0.88), and the LoA ranged from −10.63 mm (95% CI: −12.41 to −9.40) to 6.78 mm (95% CI: 5.56 to 8.57). For shuttlecock speed (Fig. 5b), the bias was −1.00 m/s (95% CI: −2.46 to 0.46), and the LoA ranged from −4.89 m/s (95% CI: −8.48 to −3.72) to 2.90 m/s (95% CI: 1.73 to 6.49).

For all four metrics, no significant correlation was found between the mean and the difference (impact time: $r = 0.013$, $p = 0.88$; medio-lateral: $r = 0.045$, $p = 0.63$; longitudinal: $r = -0.050$, $p = 0.60$; shuttlecock speed: $r = -0.022$, $p = 0.81$), indicating no proportional bias.

**4. Discussion**

This study proposes an event-camera-based method to estimate impact time, impact location, and shuttlecock speed within each trial. Agreement with the HS-based reference method was evaluated using Bland–Altman analysis for all metrics.

For impact location, the medio-lateral and longitudinal biases were 3.45 mm and −1.92 mm, respectively, indicating generally good agreement between the two methods. According to BWF regulations, the shuttlecock base diameter is 25–28 mm [13], and the string spacing of a typical racket is approximately 10 mm. In this context, the observed differences are likely small enough for practical trend analysis in performance and equipment evaluation. It should be noted that the spatial resolution of the rear-view measurement was approximately 1.86 mm/pixel, and image-resolution-related uncertainty is inherent in both methods.

The LoA width, however, ranged from several millimeters to more than ten millimeters, indicating that non-negligible variability exists at the individual-trial level. Possible contributing factors include shuttlecock deformation during contact, variation in the contact point within the contact duration, the racket face not being a perfect ellipse, and resolution-related or manual/semi-automatic annotation errors in the HS reference. Nevertheless, these differences are unlikely to affect practical trend analysis in performance and equipment evaluation and are considered acceptable for the intended application.

For impact time, the bias of 1.84 ms and LoA of −0.84 to 4.52 ms were small; however, given that the reported contact duration between the shuttlecock and the racket is approximately 1.0–1.4 ms [1, 4], the difference is not entirely negligible. The proposed method defines impact time from the inflection point of the shuttlecock trajectory in the event data, whereas the HS reference defines it as the frame at which the shuttlecock completely tip separates from the string bed. Because these two definitions do not strictly coincide, the observed bias may include not only estimation error but also the difference in reference point within the impact phenomenon.

The shuttlecock speed bias was −1.00 m/s, corresponding to approximately 2% of the mean speed of 52.64 m/s obtained by the HS reference. The LoA of −4.89 to 2.90 m/s indicates that differences of several meters per second can occur in individual trials. Shuttlecock speed decays rapidly with flight distance after a smash due to high aerodynamic drag [5]. Furthermore, immediately after impact the shuttlecock undergoes posture change and flipping, during which the cork tip and the center of mass travel at different speeds until the shuttlecock axis aligns with the flight direction [4, 5]. In addition, given the lateral-view spatial resolution of 3.47 mm/pixel, a one-pixel positional error over the 2 ms measurement interval corresponds to a speed difference of approximately 1.74 m/s, suggesting that image-resolution-related uncertainty contributed to the observed differences. Considering these factors, the observed speed differences are unlikely to significantly influence the evaluation of smash speed in practical settings, and the proposed method is considered useful for identifying trends in shuttlecock speed.

Across all metrics, no significant correlation was observed between the mean and the difference, confirming the absence of proportional bias. The proposed method can therefore serve as a stable measurement tool for evaluating smash performance, which varies widely across skill levels and playing styles. Moreover, because the method associates impact time, impact location, and shuttlecock speed within the same trial, it enables integrative analysis beyond the evaluation of individual metrics.

This study has several limitations. First, although measurements were feasible in a standard gymnasium environment, specific appearance conditions were required: a white racket frame, black strings, and a dark curtain background (Fig. 1c), and the discrimination algorithm was tailored to these conditions. Generalization to diverse racket frame colors, string colors, and background conditions remains a challenge. Future work should improve the discrimination algorithm to accommodate a wider range of appearance conditions.

Second, shuttlecock speed was estimated from a two-dimensional lateral view; projection errors arise when the out-of-plane speed component is large. In this study, smash direction was controlled within a limited range to mitigate this effect, but speed estimates may be affected when the shot direction varies substantially. Establishing a protocol that constrains the shot direction, or combining multiple viewpoints, would be beneficial in future measurements.

The event data around impact also contain racket motion information (Fig. 2d), suggesting that extraction of racket head speed and shaft angle is feasible. Expanding the set of measurable parameters would provide a more comprehensive basis for evaluating the racket–shuttlecock interaction. In addition, the automated, sequential processing capability of the proposed method is well suited for adaptive experimental design, in which the measurement results of preceding trials dynamically inform subsequent experimental conditions, thereby improving the efficiency of racket evaluation and player feedback.

## 5. Conclusion

This study proposed a measurement method using event cameras to estimate impact time, impact location, and shuttlecock speed in an integrated manner within the same trial during badminton smashes. The proposed method achieved automated estimation in the majority of trials and showed small mean differences compared with the HS-based reference method, although non-negligible variability was observed at the individual-trial level. An important advantage of the proposed method is its ability to associate impact time, impact location, and shuttlecock speed within the same trial.

These results suggest that the method can serve as a useful measurement tool for performance and equipment evaluation in badminton.

**Acknowledgements**

The authors acknowledge all participants for their commitment and engagement.

**Statements and Declarations**

**Competing interests**

The authors are employees of Mizuno Corporation. Mizuno Corporation has filed patent applications and commercial interests related to measurement technologies described in this manuscript. The authors declare no other financial or non-financial competing interests.

**Funding**

No external funding was received for this study.

**Ethics**

Ethics approval was granted by MIZUNO Corporation Research Ethics Committee (Ref. No. 20260408-475).

**Consent**

Written informed consent was obtained from all participants.

**Data Availability**

The datasets generated and/or analyzed during the current study are available from the corresponding author on reasonable request.

**Author contributions**

Y.W. conceived and designed the study, led the experimental planning and execution, performed data processing, and integrated all sections into the final manuscript.

Y.K. contributed to the overall algorithmic concept and methodological discussions, developed and implemented core software components, and drafted relevant sections.

R.Y. implemented the software for impact time estimation and shuttlecock speed estimation and drafted relevant sections.
K.I. developed and implemented the software for camera synchronization and impact-location estimation, integrated the processing pipeline, and drafted relevant sections.
S.H. conducted experiments and performed data processing and statistical analysis and drafted relevant sections.
All authors contributed to critical revision and discussion, approved the final manuscript, and agreed to be accountable for all aspects of the work.

**References**

1. Towler H, Mitchell SR, King MA (2023) Effects of racket moment of inertia on racket head speed, impact location and shuttlecock speed during the badminton smash. Sci Rep 13:14060. https://doi.org/10.1038/s41598-023-37108-x
2. McErlain-Naylor SA, Towler H, Afzal IA, Felton PJ, Hiley MJ, King MA (2020) Effect of racket-shuttlecock impact location on shot outcome for badminton smashes by elite players. J Sports Sci 38(21):2471–2478. https://doi.org/10.1080/02640414.2020.1792132
3. Ramasamy Y, Yeap MW, Towler H, King M (2025) Intra-individual variation in the jump smash for elite Malaysian male badminton players. Appl Sci 15(2):844. https://doi.org/10.3390/app15020844
4. Cohen C, Darbois Texier B, Quéré D, Clanet C (2015) The physics of badminton. New J Phys 17:063001. https://doi.org/10.1088/1367-2630/17/6/063001
5. Collet E (2026) Shuttlecock velocity decay after smash and slice shots in badminton. Phys Scr 101:125007. https://doi.org/10.1088/1402-4896/ae5361

6. Gallego G, Delbruck T, Orchard G, Bartolozzi C, Taba B, Censi A et al (2022) Event-based vision: a survey. IEEE Trans Pattern Anal Mach Intell 44(1):154–180. https://doi.org/10.1109/TPAMI.2020.3008413

7. Gossard T, Krismer J, Ziegler A, Tebbe J, Zell A (2024) Table tennis ball spin estimation with an event camera. In: IEEE/CVF Conference on Computer Vision and Pattern Recognition Workshops (CVPRW), pp 3347–3356. https://doi.org/10.1109/CVPRW63382.2024.00339

8. Nakabayashi T, Higa K, Yamaguchi M, Fujiwara R, Saito H (2024) Event-based ball spin estimation in sports. In: IEEE/CVF Conference on Computer Vision and Pattern Recognition Workshops (CVPRW), pp 3367–3375. https://doi.org/10.1109/CVPRW63382.2024.00341

9. Kase Y, Ishibe K, Yasuda R, Washida Y, Hashimoto S (2025) Locating tennis ball impact on the racket in real time using an event camera. In: Sports Analytics (ISACE 2025). Lecture Notes in Computer Science, vol 15925. Springer, Cham, pp 99–115. https://doi.org/10.1007/978-3-032-06167-6_8

10. Yasuda R, Hashimoto S, Kase Y, Ishibe K, Washida Y (2025) Estimating shuttlecock speed in badminton smashes with an event camera. Sports Informatics and Technology 2025, Session ID B-3-1. https://doi.org/10.82713/sit.2025.0_B-3-1 (in Japanese)
11. Giavarina D (2015) Understanding Bland Altman analysis. Biochem Med (Zagreb) 25(2):141–151. https://doi.org/10.11613/BM.2015.015
12. Zou GY (2013) Confidence interval estimation for the Bland–Altman limits of agreement with multiple observations per individual. Stat Methods Med Res 22(6):630–642. https://doi.org/10.1177/0962280211402548
13. Badminton World Federation (BWF). Statutes: Section 4.1 – Laws of Badminton. Available at: https://corporate.bwfbadminton.com/statutes/#1513733461252-a16ae05d-1fc9(Accessed: 18 May 2026).

Fig. 1 Overview of the proposed system. (a) Experimental setup in an indoor gymnasium, showing sensor placement and the measurement environment. (b) Processing pipeline for estimating the swing interval, impact time, shuttlecock tracking, and impact location. (c) Racket used in the experiment, featuring a white frame and black strings.

Fig. 2 Swing interval detection and impact time estimation based on event rate. (a) Mean event rate over time. (b) Variance of the event rate over time. Solid lines indicate the respective statistics; dashed lines indicate the detected swing interval. (c) Event distribution within the swing interval in x–y–t space (white: positive polarity; blue: negative polarity). (d) x–y projection of the swing interval. (e) x–t projection, where the trajectory inflection point was detected as the impact time.

Fig. 3 Impact location estimation from rear-view event images. (a) Extraction of the region of interest from the rear-view event image. (b) Extraction of positive-polarity regions and separation of the racket-face and shaft regions. (c) Ellipse fitting to the racket face, definition of the racket-face coordinate system, and computation of the impact location; the inset shows a cropped view illustrating the u–v coordinate axes.

Fig. 4 Comparison of representative results from a single trial obtained by the high-speed-camera-based reference method (HS) and the proposed event-camera-based method (EV). (a) Impact location estimated by the HS method. (b) Shuttlecock speed estimated by the HS method. (c) Impact location estimated by the EV method. (d) Shuttlecock speed estimated by the EV method.

Fig. 5 Bland–Altman plots comparing the estimates obtained by the proposed event-

camera-based method (EV) and the high-speed camera-based reference method (HS). (a) Impact time. (b) Shuttlecock speed. (c) Medio-lateral impact location. (d) Longitudinal impact location. Differences were calculated as EV − HS. Solid lines indicate the bias (mean difference); dashed lines indicate the 95% limits of agreement (mean difference ± 1.96 SD).

Fig. 1

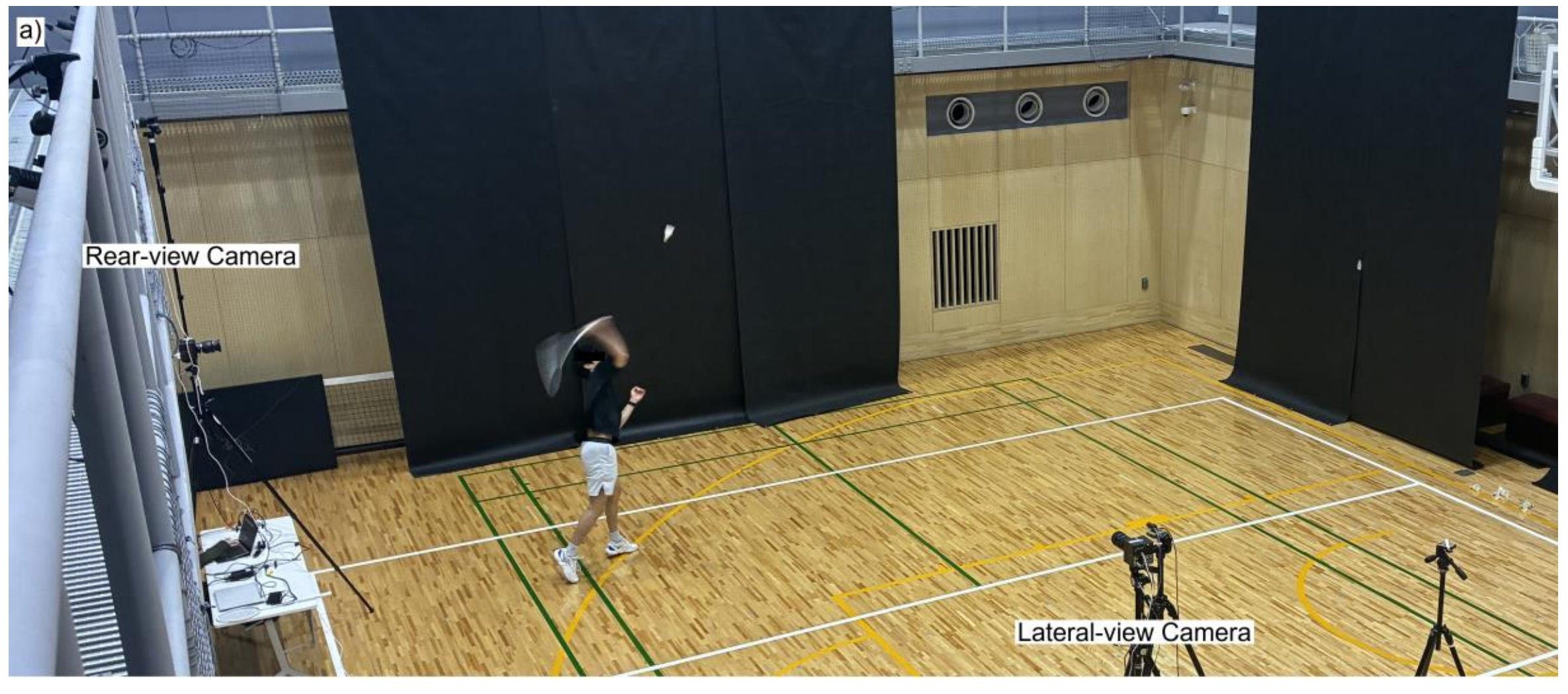
a)
Rear-view Camera
Lateral-view Camera

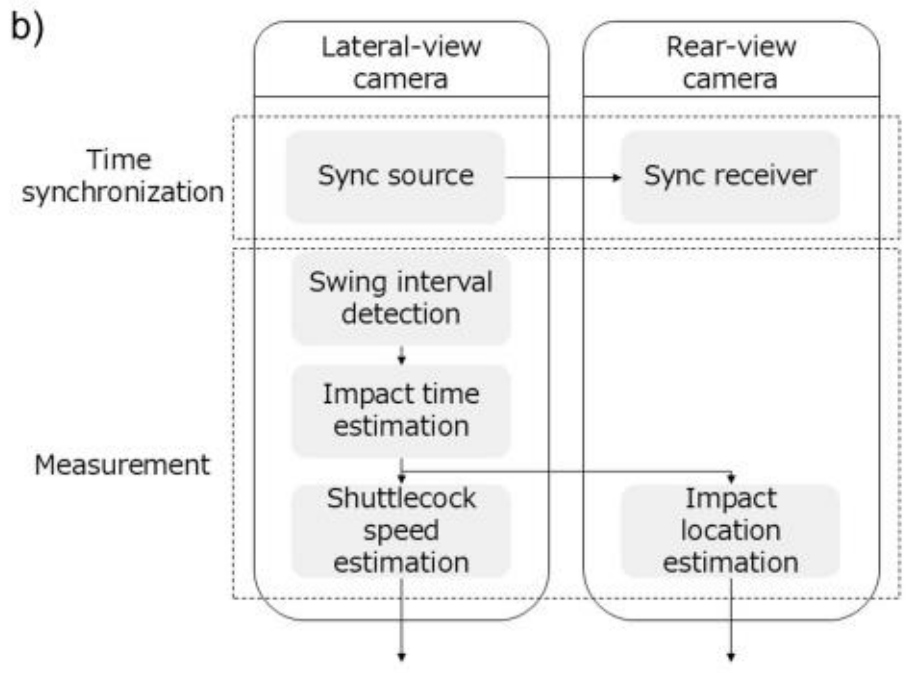
b)
Lateral-view camera
Rear-view camera
Time synchronization
Sync source
Sync receiver
Swing interval detection
Impact time estimation
Measurement
Shuttlecock speed estimation
Impact location estimation

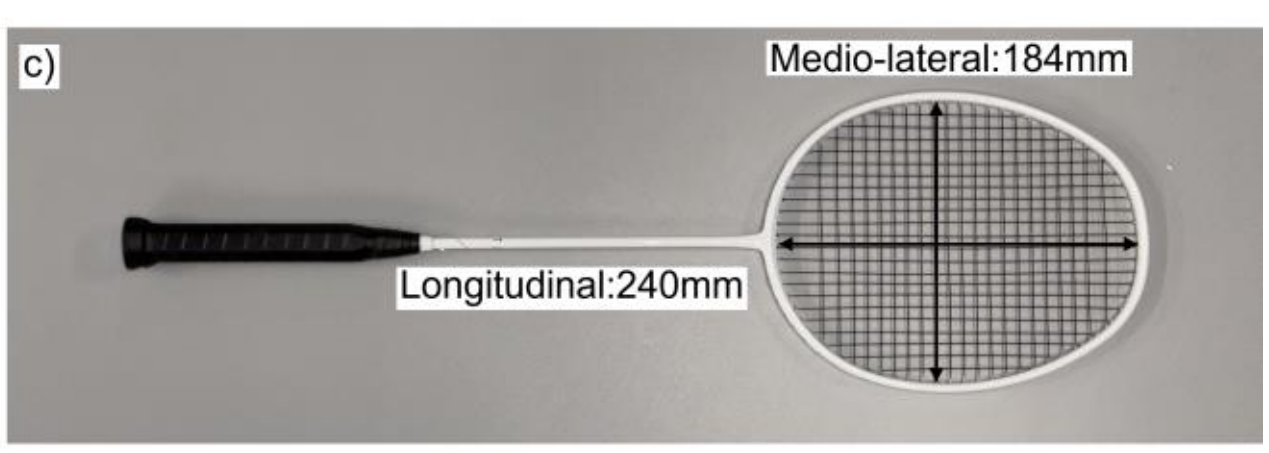
c)
Medio-lateral:184mm
Longitudinal:240mm

Fig. 2

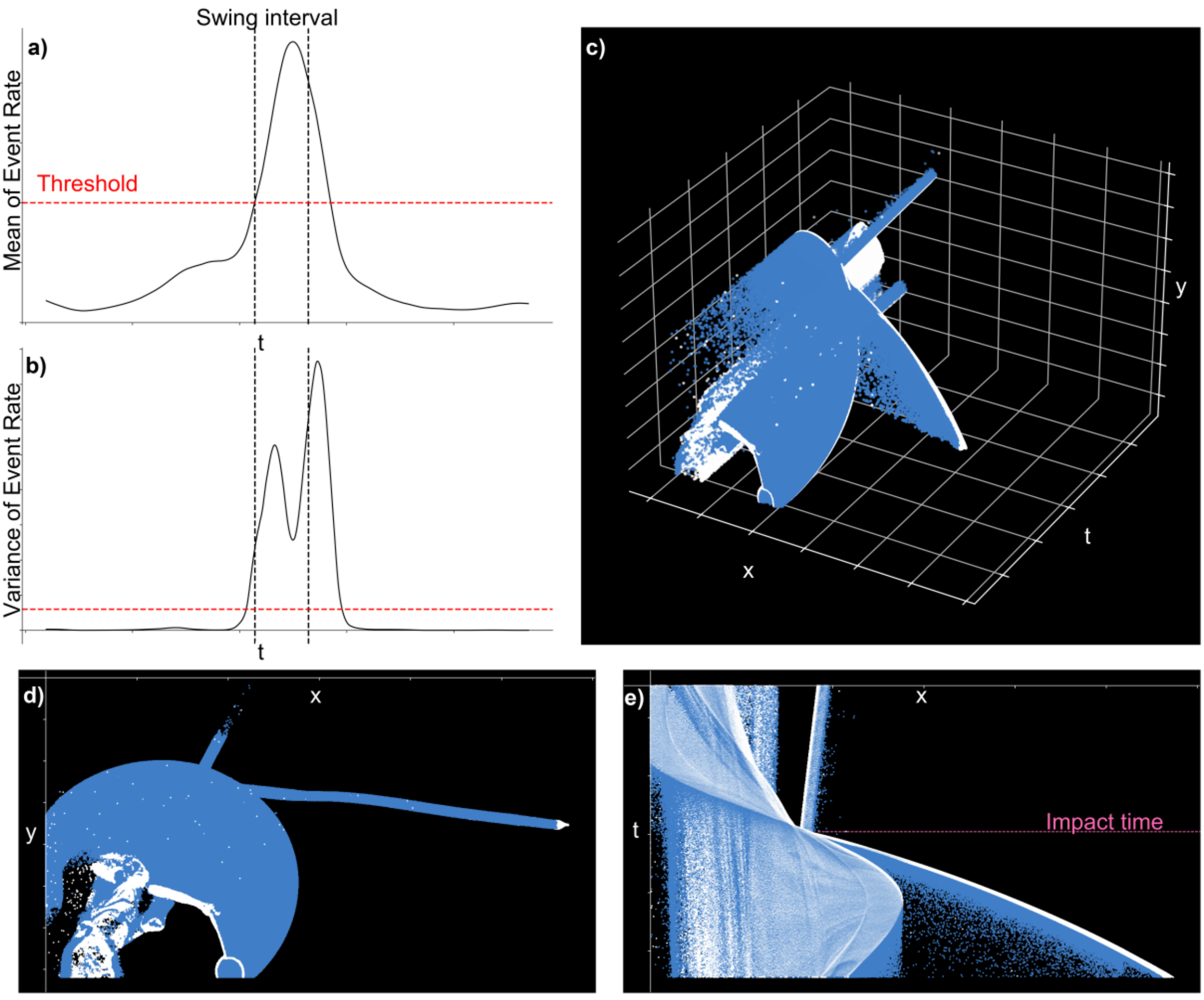

Swing interval
a)
Mean of Event Rate
Threshold
t
b)
Variance of Event Rate
t
c)
y
t
x
d)
x
y
e)
x
t
Impact time

Fig. 3

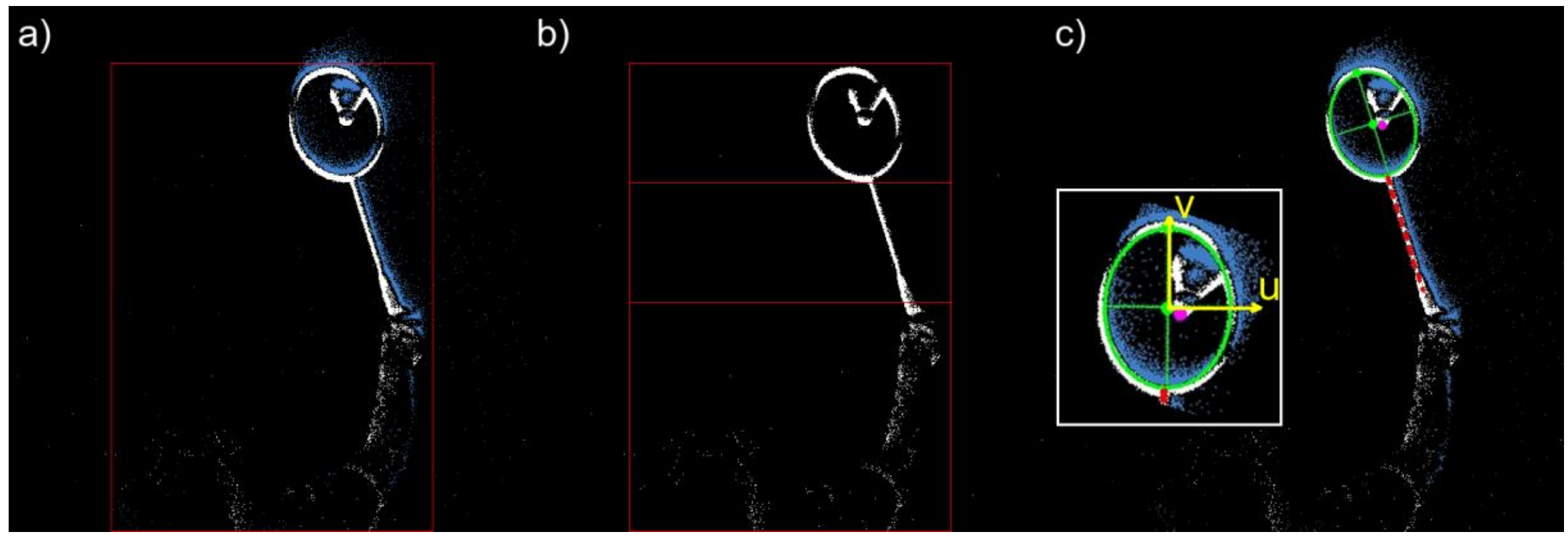

Fig. 4

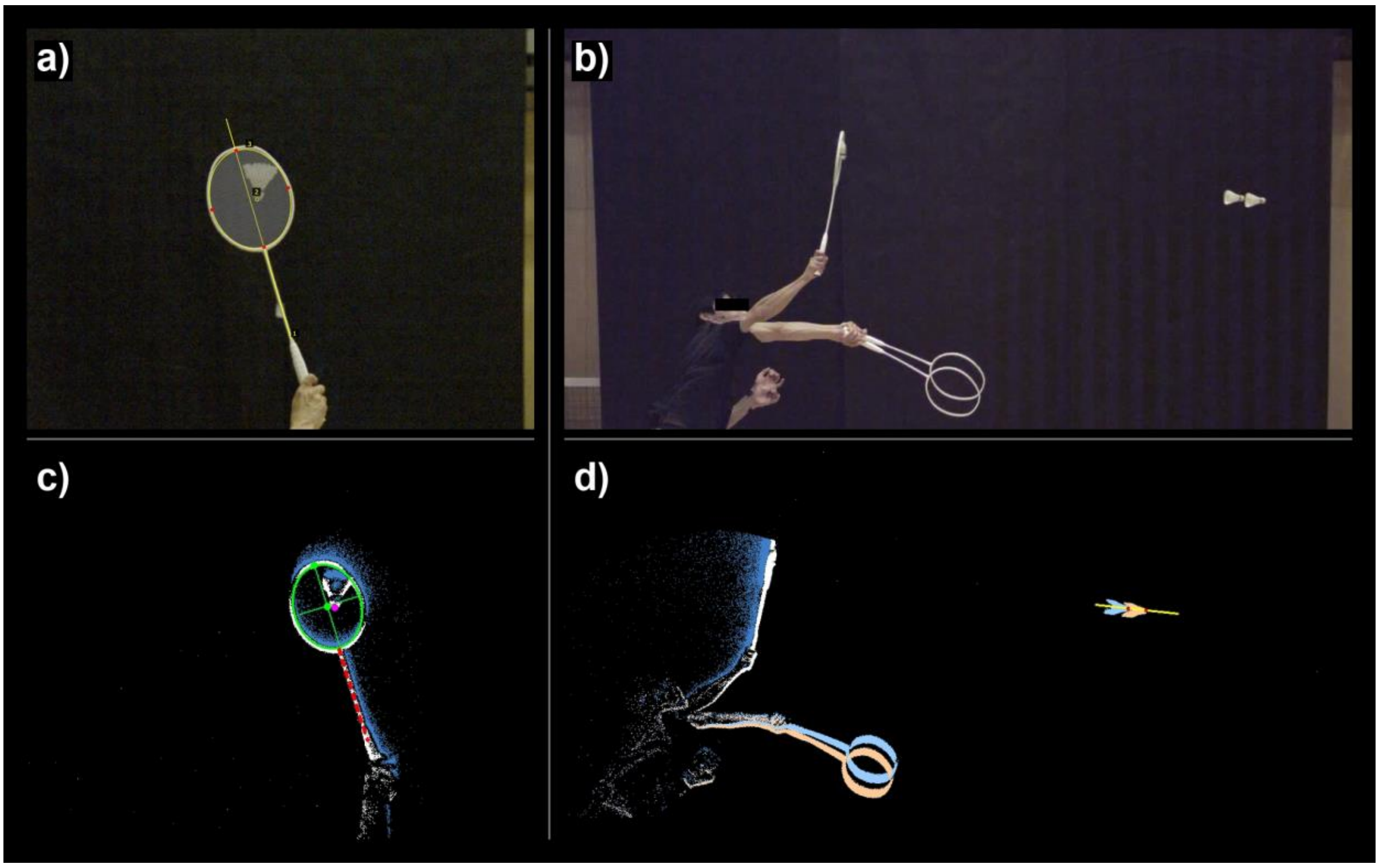

Fig.5

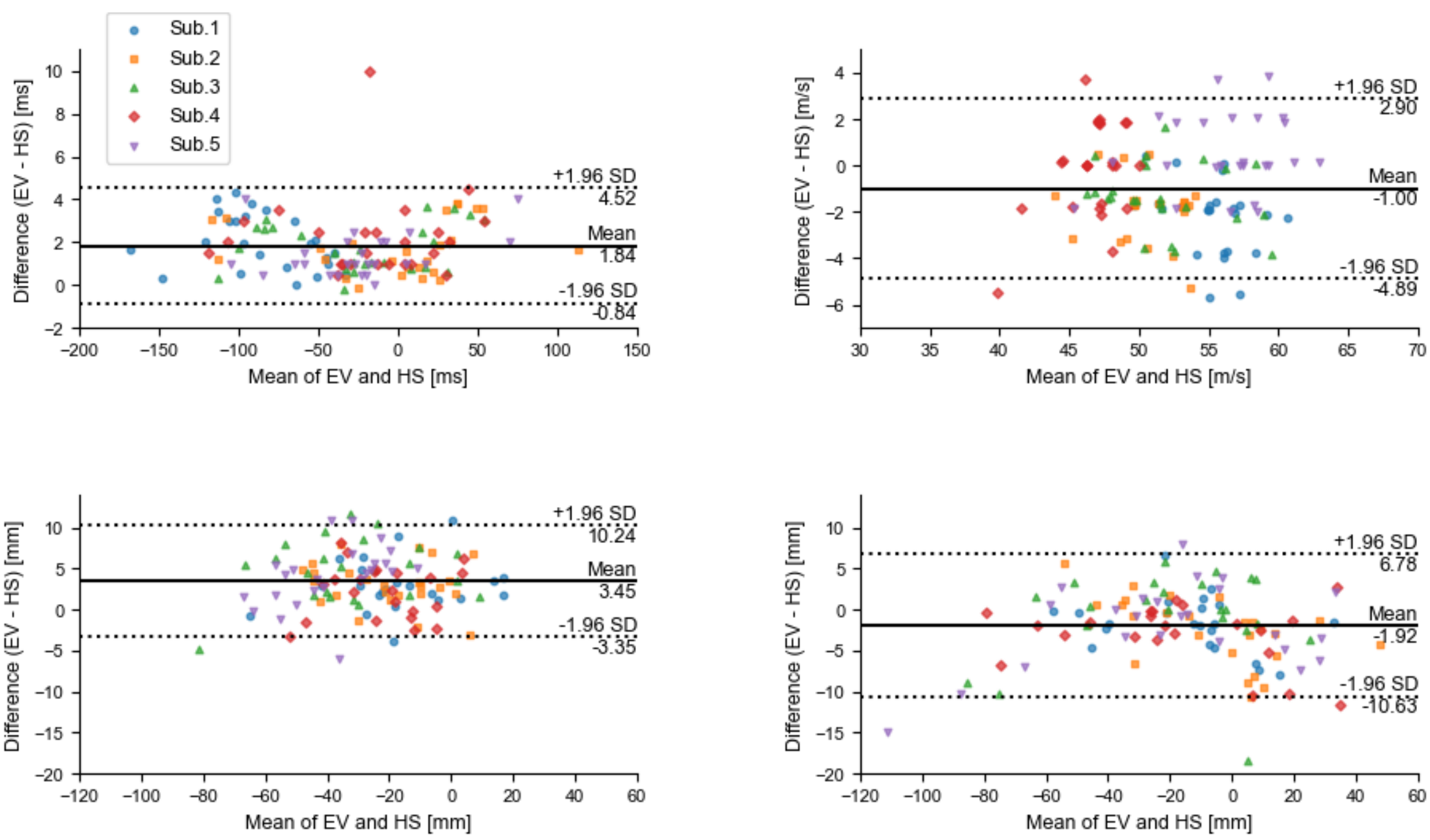

Sub.1
Sub.2
Sub.3
Sub.4
Sub.5
Difference (EV - HS) [ms]
Mean of EV and HS [ms]
+1.96 SD
4.52
Mean
1.84
-1.96 SD
-0.84
Difference (EV - HS) [m/s]
Mean of EV and HS [m/s]
+1.96 SD
2.90
Mean
-1.00
-1.96 SD
-4.89
Difference (EV - HS) [mm]
Mean of EV and HS [mm]
+1.96 SD
10.24
Mean
3.45
-1.96 SD
-3.35
Difference (EV - HS) [mm]
Mean of EV and HS [mm]
+1.96 SD
6.78
Mean
-1.92
-1.96 SD
-10.63